\documentclass[10pt,twocolumn,letterpaper]{article}

\usepackage{cvpr}      

\definecolor{cvprblue}{rgb}{0.21,0.49,0.74}
\usepackage{float}
\usepackage{graphicx}
\usepackage{subcaption}
\usepackage{multirow}
\usepackage{adjustbox}
\usepackage[accsupp]{axessibility} 
\usepackage[pagebackref,breaklinks,colorlinks,allcolors=cvprblue]{hyperref}

\def\paperID{108} 
\def\confName{CVPR}
\def\confYear{2025}

\title{Enhanced Semantic Extraction and Guidance for UGC Image Super Resolution}

\author{
    Yiwen Wang$^{1}$ \quad 
    Ying Liang$^{1}$ \quad 
    Yuxuan Zhang$^{1}$ \quad
    Xinning Chai$^{1}$ \quad
    Zhengxue Cheng$^{1}$\thanks{ Corresponding author} \\ 
    Yingsheng Qin$^{2}$ \quad
    Yucai Yang$^{2}$ \quad 
    Rong Xie$^{1}$ \quad
    Li Song$^{1}$ \\
    $^1$Shanghai Jiao Tong University, China \quad $^2$Transsion, China \\
    {\tt\small \{evonwang, forest726, 67keudyhsi, chaixinning, zxcheng, xierong, song\_li\}@sjtu.edu.cn} \\
    {\tt\small \{yingsheng.qin, yucai.yang\}@transsion.com}
}

\begin{document}

\maketitle
\begin{abstract}
Due to the disparity between real-world degradations in user-generated content(UGC) images and synthetic degradations, traditional super-resolution methods struggle to generalize effectively, necessitating a more robust approach to model real-world distortions. In this paper, we propose a novel approach to UGC image super-resolution by integrating semantic guidance into a diffusion framework. Our method addresses the inconsistency between degradations in wild and synthetic datasets by separately simulating the degradation processes on the LSDIR dataset and combining them with the official paired training set. Furthermore, we enhance degradation removal and detail generation by incorporating a pretrained semantic extraction model (SAM2) and fine-tuning key hyperparameters for improved perceptual fidelity. Extensive experiments demonstrate the superiority of our approach against state-of-the-art methods. Additionally, the proposed model won second place in the CVPR NTIRE 2025 Short-form UGC Image Super-Resolution Challenge (Report \cite{ntire2025shortugc,ntire2025shortugc_data}), further validating its effectiveness. The code is available at \url{https://github.com/Moonsofang/NTIRE-2025-SRlab}.
\end{abstract}    
\section{Introduction}
\label{sec:intro}
\begin{figure}[t]
    \centering
    \includegraphics[width=0.9\linewidth]{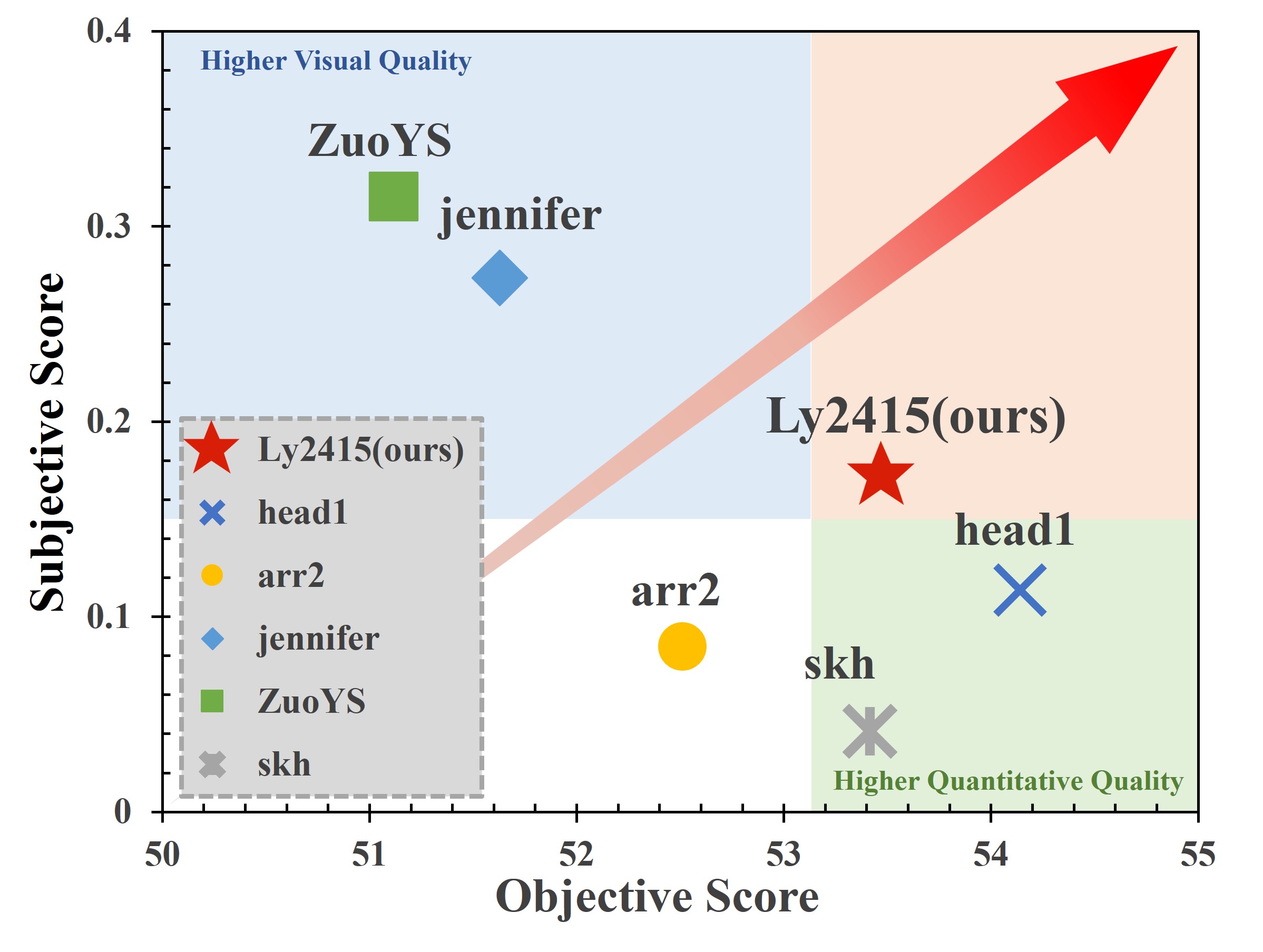}
    \caption{Objective and subjective results of NTIRE 2025 Short-form UGC Image SR Challenge. The top six methods are included. The horizontal axis represents the objective score, which is computed as $\text{Score}=\text{PSNR}+10\times \text{SSIM}-10\times \text{LPIPS}+0.1\times \text{MUSIQ}+10\times \text{ManIQA}+10\times \text{CLIPIQA}$. The vertical axis represents the subjective score calculated by five experts. All results above are provided by the competition organizer.}
    \label{fig:enter-label}
\end{figure}

Single Image Super-Resolution (SISR) is a fundamental task in computer vision that focuses on reconstructing high-resolution (HR) images from their low-resolution (LR) counterparts. The primary goal of SISR is to recover fine-grained details and high-frequency textures that are lost during the downsampling process, ultimately enhancing the perceptual quality and fidelity of the upscaled images.

Early deep learning-based approaches predominantly relied on convolutional neural networks (CNNs)  \cite{SRCNN,EDSR,VDSR,RCAN,DRCN,RDN}, which leverage hierarchical feature extraction to learn LR-to-HR mappings.   While these methods achieve impressive PSNR, they often produce overly smooth textures due to their reliance on pixel-wise loss functions, failing to capture high-frequency details essential for perceptual realism. By adversarial training, GAN-based methods  \cite{LDL,RealSR,SRGAN,ESRGAN,BSRGAN,Real-ESRGAN} synthesize visually plausible textures, significantly improving perceptual quality. However, their tendency to hallucinate visually plausible yet semantically inconsistent details remains a critical limitation for faithful super-resolution.

Unlike GANs, Diffusion models(DMs) \cite{DDPMs} iteratively refine images through a Markov chain-based denoising process. This progressive nature allows for finer control over the trade-off between distortion and perceptual quality, making them particularly well-suited for high-fidelity image generation. Additionally, their inherent stochasticity enables the generation of diverse high-resolution (HR) samples from a single low-resolution (LR) input, a distinct advantage over traditional deterministic approaches such as CNNs and GANs, which typically produce a single fixed output. Recently, a growing number of diffusion-based super-resolution (SR) methods  \cite{IDM,DiffIR,SeeSR,diffbir,resshift,supir,pasd,pisa,faithdiff,invsr,xpsr} have emerged, further demonstrating the effectiveness of this framework in enhancing image quality and restoring fine details. 

However, despite their promising capabilities, diffusion models still face several critical challenges when applied to real-world super-resolution tasks.

First, most existing methods rely on synthetically generated low-resolution (LR) and high-resolution training pairs, where degradation processes, such as bicubic downsampling, fail to accurately replicate the complex and heterogeneous degradations found in real-world images. This discrepancy creates a domain gap, making it difficult for models trained on synthetic data to generalize effectively to real-world scenarios. User-generated content (UGC) images, for instance, often suffer from diverse and unpredictable degradation patterns, including noise, compression artifacts, motion blur, and varying lighting conditions, further complicating the super-resolution process. As a result, models optimized for artificially degraded datasets may struggle to restore authentic details when applied to in-the-wild data.

Second, while diffusion models excel at generating plausible high-frequency details, they often struggle to maintain a delicate balance between degradation removal and semantic fidelity. In many cases, the model may introduce unnatural artifacts or fail to preserve the structural coherence of objects, leading to inconsistencies in the restored images. This issue is particularly pronounced in cases where extreme degradation has erased fine-grained textures, making it challenging for the model to reconstruct missing information in a visually meaningful way.

To address these challenges, we propose a diffusion-based super-resolution (SR) framework that integrates realistic degradation modeling, semantic-aware refinement, and perceptual optimization. Our method aims to bridge the gap between synthetic and real-world wild degradation scenarios by constructing a more representative training dataset and incorporating advanced architectural components to enhance restoration quality. Specifically, we introduce a training strategy that combines the synthetic training set provided by the competition with the LSDIR dataset \cite{LSDIR}. To better simulate real-world conditions, we further process the LSDIR dataset by applying controlled degradation, ensuring that the model learns to handle diverse degradation patterns effectively.

Our approach leverages the strong generative prior of Stable Diffusion \cite{sd}, which provides a powerful latent space for high-quality image synthesis. To further enhance reconstruction fidelity, we integrate ControlNet \cite{controlnet}, which enables precise spatial conditioning and improves structural detail preservation. Additionally, we incorporate a semantic-aware module to refine both structural and contextual information, ensuring that the generated high-resolution images maintain both perceptual quality and semantic coherence. Specifically, we leverage the semantic extraction capabilities of SAM2 \cite{SAM2} to recover fine-grained details by embedding high-level semantic information into the latent space. This integration allows our framework to strike a balance between effective degradation removal and realistic detail generation, making it more robust to diverse real-world degradations while producing visually coherent and semantically meaningful reconstructions.

Our contributions are as follows.
\begin{itemize}
\item To address the inconsistency between degradations in wild and synthetic datasets, we simulate both synthetic and wild degradation processes on LSDIR separately.  The resulting datasets are then combined with the officially provided paired training set to form the final training dataset.

\item To enhance degradation removal and semantic detail generation, we incorporate the pretrained semantic extraction model SAM, which helps refine the structural and contextual understanding of the input data.

\item To better control the visual quality of the generated results, we fine-tune specific hyperparameters of the model, optimizing its performance for improved perceptual fidelity.
\end{itemize}

\section{Related Works}
\label{sec:related}
In recent years, the advancement of deep learning technology has significantly propelled progress in Single Image Super-Resolution (SISR). Based on a variety of deep learning networks, the capability of SISR models to improve the quality of low-resolution images has been greatly enhanced. The commonly employed architectures in this domain includes Convolutional Neural Networks (CNNs), Generative Adversarial Networks (GAN), Transformer, and Diffusion model \cite{review}.

\subsection{CNN-Based SISR}
CNN is a subclass of feedforward neural networks that incorporate convolutional operation. Its characteristics of local connectivity and weight sharing endow it with powerful feature-learning capabilities. 

Notable examples of CNN-based single-image super-resolution (SISR) models include SRCNN \cite{SRCNN} (the first deep learning-based SISR model), BSRN \cite{BSRN} (won the first place in the NTIRE 2022 Efficient SR (ER) track), and CVANet \cite{CVANet}. An important research direction for CNN-based SISR models is enhancing network depth and width to improve the model's ability to capture fine image details.

\subsection{GAN-Based SISR}
Generative Adversarial Networks (GANs) consist of a generator and a discriminator. Through adversarial training, GANs effectively learn the underlying distribution of images, e.g., LDL \cite{LDL} and RealSR \cite{RealSR}, are GAN-based SISR models. GAN-based SISR models have demonstrated significant advantages in terms of perceptual quality and realistic visual effects. However, the large number of network parameters leads to unstable training and relatively long inerence time. Enhancing reconstruction capabilities of GAN-based models while designing lightweight networks to stabilize the training process has become a key research direction.

\subsection{Transformer-Based SISR}
The Transformer architecture, originally designed for natural language processing, is a neural network framework that relies entirely on self-attention mechanisms to capture global dependencies between inputs and outputs. Compared to CNN and RNN, self-attention enables better modeling of long-range relationships while mitigating the vanishing gradient problem. Additionally, Transformers support parallel computation, significantly improving network efficiency.  

A Transformer network consists of an encoder-decoder structure, with multi-head self-attention as its core component. Models such as GRL \cite{GRL}, HIPA \cite{HIPA}, and ESRT \cite{ESRT} are based on Transformer architectures. Despite their advantages, Transformer-based SISR models still face challenges such as high computational costs and artifacts that affect perceptual quality. Addressing these issues remains an important research direction for improving the application of Transformers in image super-resolution.

\subsection{Diffusion-Based SISR}
In recent years, diffusion models have emerged as a powerful approach based on the principle of destruction and reconstruction, incorporating both forward and backward diffusion processes. Notable diffusion-based SISR models include IDM \cite{IDM}, DiffIR \cite{DiffIR}, and SeeSR \cite{SeeSR}, etc.

Most existing diffusion-based SISR models adopt the UNet architecture from denoising diffusion probabilistic models (DDPMs) \cite{DDPMs}, while a few employ entirely novel structures, such as Transformer-integrated diffusion models. Designing a unified and foundational diffusion model architecture is a promising research direction in the SISR field.  

However, the introduction of stochastic noise in the backward diffusion process often results in unstable outputs. Therefore, developing a stable and efficient diffusion-based SISR model is a forward-looking research challenge, aiming to balance reconstruction quality with computational efficiency.

\section{Method}
\label{sec:method}

\subsection{Preliminaries}
\noindent\textbf{Diffusion Model}
Diffusion models \cite{deep,DDPMs} are a class of generative models that achieve the goal of generating target data samples from Gaussian noise $\mathbf{x}_T \sim \mathcal{N}(0,1)$. Diffusion models are based on a two-step process: a forward diffusion process and a reverse denoising process.

In the forward diffusion process, a data sample $\mathbf{x}_0$ is gradually perturbed by adding Gaussian noise over $T$ timesteps:
\begin{equation}
    q(\mathbf{x}_t | \mathbf{x}_{t-1}) = \mathcal{N}(\mathbf{x}_t; \sqrt{1 - \beta_t} \mathbf{x}_{t-1}, \beta_t \mathbf{I}),
\end{equation}
where $\beta_t$ is a variance schedule controlling the noise level at each timestep, $\mathbf{x}_t$ is the noised image at timestep $t$.

In the reverse process, a neural network $\epsilon_\theta(\mathbf{x}_t, t)$ is trained to predict the added noise, enabling the generation of new samples by denoising a random Gaussian sample $\mathbf{x}_T$:
\begin{equation}
    p_\theta(\mathbf{x}_{t-1} | \mathbf{x}_t) = \mathcal{N}(\mathbf{x}_{t-1}; \mu_\theta(\mathbf{x}_t, t), \Sigma_\theta(\mathbf{x}_t, t)).
\end{equation}
where $\mu_\theta(\mathbf{x}_t, t)$ is typically parameterized using the predicted noise $\epsilon_\theta(\mathbf{x}_t, t)$ and the noise schedule.

In practice, denoising models adopt a denoising network $\epsilon_\theta(\mathbf{x}_t, t)$ to estimate noise $\epsilon$. During training, the network parameters $\theta$ of the denoising network $\epsilon_\theta$ are optimized by minimizing a loss function
\begin{equation}
\mathcal{L}=\mathbb{E}_{\text{x}_0,t,\epsilon} \left\| \epsilon - \epsilon_{\theta}(\text{x}_t, t) \right\|_2^2
\end{equation}

Recent advancements, such as classifier-free guidance \cite{cfg}, have further enhanced the capability of diffusion models, making them state-of-the-art in various generative modeling tasks.

\subsection{Overview}
\begin{figure*}[t]
  \centering
  \includegraphics[width=0.85\linewidth]{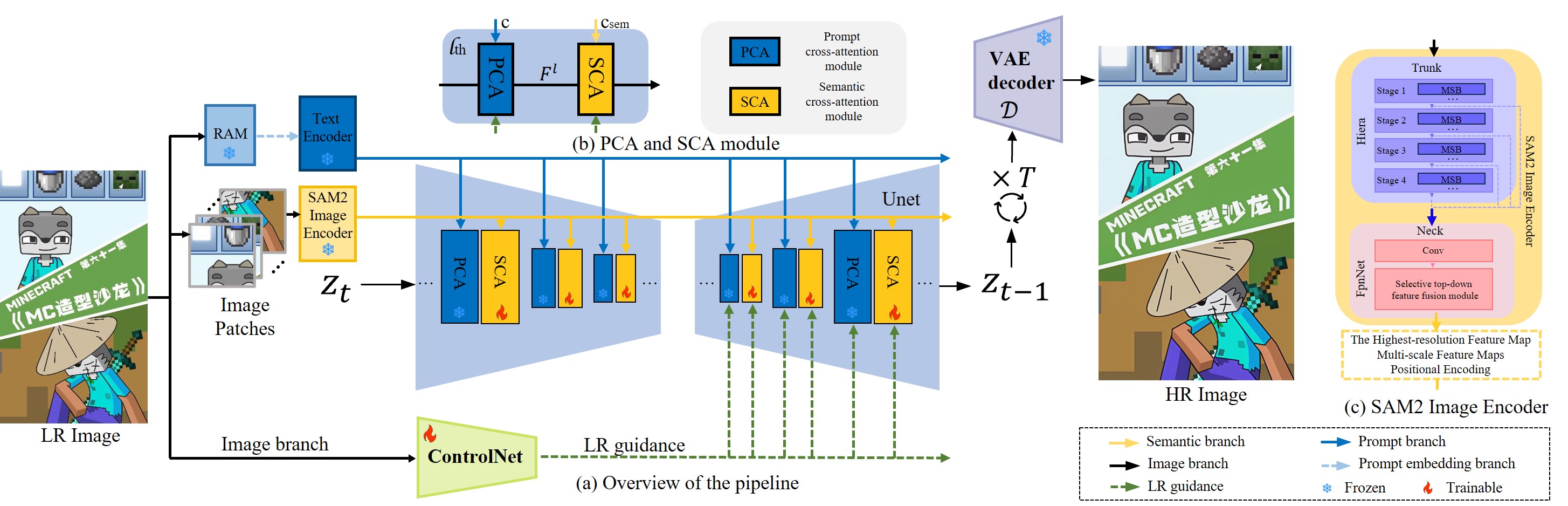}
  \caption{(a) Overview of our proposed method. Our approach builds upon the diffusion framework, incorporating a mechanism to enforce structural consistency and preserve fidelity to the original LR image. Additionally, we leverage SAM2 for semantic-guided refinement, extracting high-level semantic embeddings to enhance adaptability to diverse degradation conditions. (b) Architecture of the PCA and SCA module. (c) Architecture of the SAM2 image encoder. The encoder comprises a trunk and a neck. The trunk extracts multi-scale features from the low-resolution image through four stages with varying numbers of MSB(Multi-Scale Block) layers. The neck applies convolutions at all scales and performs top-down feature fusion on low-resolution features. The output includes refined multi-scale feature maps and corresponding positional encodings.}
  \label{fig:pipe}
\end{figure*}

Our proposed method is built upon the diffusion framework. The overall pipeline of our approach is illustrated in \cref{fig:pipe}. Given a low-quality input image $I_{\text{LR}}$, we first encode it into a latent representation using VAE encoder \cite{vae}.
To progressively refine the latent representation, we employ a Denoising U-Net, which iteratively removes noise through a sequence of denoising steps. Additionally, we incorporate ControlNet architecture to enforce structural consistency with the low-resolution (LR) input. ControlNet enables precise conditioning, ensuring that the super-resolved output maintains fidelity to the original image while benefiting from the generative power of diffusion models.  

A key challenge in real-world super-resolution tasks is handling images with varying degrees of degradation and different scaling factors, particularly when dealing with both synthetic and wild test sets. To address this, we integrate the Segment Anything Model 2 (SAM2) \cite{SAM2} into our framework. SAM2 is utilized to extract high-level semantic embeddings from the input images, providing additional contextual information that aids in the denoising and reconstruction process. These semantic embeddings help our model adapt to diverse image conditions, improving robustness across different datasets. The enriched latent representation, augmented with semantic information, is then iteratively refined by the Denoising U-Net over $T$ denoising steps.  

During training, we minimize a denoising objective:  
\begin{equation}
\mathcal{L}=\mathbb{E}_{\text{x}_0,\text{x}_{LR},t,c,c_{sem},\epsilon} \left\| \epsilon - \epsilon_{\theta}(\text{x}_t, \text{x}_{LR}, t, c, c_{sem}) \right\|_2^2
\end{equation}
where $\text{x}_{LR}$ represents the low-resolution (LR) latent, $c$ denotes the tag prompt, and $c_{sem}$ is the semantic embedding. The noise estimation network $\epsilon_{\theta}$ is responsible for predicting the noise $\epsilon \sim \mathcal{N}(0,\mathbf{I})$. 

\subsection{Semantic-Aware Module}
To ensure robustness across inputs $I_{\text{LR}}$ with different levels of degradation, as illustrated in \cref{fig:pipe}(b), we introduce the Semantic-Aware Module, which extracts high-level semantic embeddings from each frame using a pre-trained SAM2 model. Specifically, as formulated in \cref{eq:semantic_eq1}, the input images are processed through the frozen image semantic extractor to obtain their corresponding semantic representations $c_{sem}$:  

\begin{equation}
\label{eq:semantic_eq1}
\begin{aligned}
c_{sem} = \texttt{Semantic Extractor}(I_{\text{LR}}) \\
\end{aligned}
\end{equation}

The extracted semantic embeddings effectively capture essential structural and contextual information, even in the presence of severe degradation. These features are subsequently integrated into the denoising U-Net through a semantic attention mechanism, facilitating a more informed restoration process. In particular, at the $l$-th layer of the U-Net, we compute the query vector $Q$ from the spatial feature map $F^l$, while the key $K$ and value $V$ vectors are derived from the semantic embedding $c_{sem}$, as defined in \cref{eq:semantic_eq2}:

\begin{equation}
\label{eq:semantic_eq2}
\begin{aligned}
& Q = W_q(F^l), \quad K = W_k(c_{sem}), \quad V = W_v(c_{sem}) \\
&\texttt{Attention}(Q,K,V) = \texttt{Softmax} \left(\frac{QK^T}{\sqrt{d}}\right)V \\
\end{aligned}
\end{equation}

By incorporating semantic priors extracted from SAM2, the Semantic-Aware Module enables the U-Net to leverage high-level contextual cues, thereby improving the reconstruction of fine details and preserving structural consistency. This integration significantly enhances the model’s ability to handle complex degradations, resulting in more realistic and visually coherent super-resolved outputs.

\section{Experiment}
\label{sec:exp}
\subsection{Experimental Setup}
\noindent\textbf{Train Dataset} 
To enhance the model’s performance in short-form user-generated content (UGC) scenarios, the training dataset is constructed by integrating the synthetic UGC training set provided by the competition with the LSDIR dataset \cite{LSDIR}. By leveraging multiple data sources, the training set is designed to capture a diverse and realistic range of UGC image degradations.

The dataset is composed of three complementary components. First, in the LSDIR training set, each high-resolution image undergoes either downsampling or a degradation process with an equal probability of 50\%, ensuring a balanced distribution of degradation patterns. Second, the synthetic UGC training dataset provides additional high-resolution (HR) and low-resolution (LR) training pairs, where existing paired images are cropped into overlapping 512$\times$512 sub-images to enhance sample diversity while preserving local structural details. These cropped patches retain their original HR-LR correspondence, ensuring consistency within the dataset. Finally, to further increase degradation diversity, high-resolution images from the synthetic UGC dataset are first cropped into 512$\times$512 patches and subsequently subjected to a dedicated degradation process, simulating wild UGC distortions.


\noindent\textbf{Validation Dataset and Test Dataset} 
During the validation and testing phase, the evaluation is conducted using the official validation and test sets provided by the NTIRE 2025 Challenge on Short-form UGC Image Super-Resolution ($4\times$), ensuring a comprehensive assessment of its generalization ability in synthetic and real-world UGC scenarios.

Both the validation and test sets consist of two subsets: a synthetic dataset and a wild dataset. The synthetic dataset comprises 180 paired low-resolution (LR) and high-resolution (HR) images, where LR images are generated through 4$\times$ downsampling followed by degradation, simulating distortions commonly found in real-world UGC content. Model performance on the synthetic dataset is assessed using PSNR, SSIM, and LPIPS \cite{lpips}. The wild dataset consists of 190 unpaired real-world UGC images collected from short-video platforms. These images exhibit authentic distortions but lack corresponding HR references, as they undergo degradation without downsampling. We employ no-reference metrics, including MUSIQ \cite{musiq}, ManIQA \cite{maniqa}, CLIPIQA \cite{clipiqa}, NRQM \cite{nrqm}, HyperIQA \cite{hyperiqa}, to evaluate model's restoration capability in this setting. Additionally, we utilize the DIV2K \cite{div2k} validation dataset and apply the same degradation pipeline that was used during training to generate LR images.


\noindent\textbf{Training details}
During training, the model is optimized using the integrated training dataset.   To balance learning effectiveness with computational efficiency, training is conducted for 90,000 steps on an Nvidia RTX 3090 GPU.   The Adam optimizer is utilized with a learning rate of \(5 \times 10^{-5}\). To preserve the integrity of the pre-trained Stable Diffusion \cite{sd} parameters, they remain frozen throughout training. Optimization is applied exclusively to the ControlNet component and the semantic-aware module. The training strategy enables the model to adapt effectively to short-form UGC images while preserving the pretrained features of the diffusion backbone.

\noindent\textbf{Testing details}
During testing, we analyze three key parameters: start point, guidance scale, and positive or negative prompts to assess their impact on the model's performance across synthetic and wild datasets. The goal is to determine the optimal configuration for the final test settings.

First, we compare two initialization methods: Gaussian noise and noised low-resolution latent, assessing their impact on super-resolved image quality. Next, we examine the effect of positive and negative prompts. Additional positive prompts such as clean, high-resolution, 8K, ultra-detailed, ultra-realistic are introduced to enhance image sharpness and fidelity. Conversely, negative prompts including dotted, noise, blur, low-resolution, smooth, unrealistic physics, unnatural shadows are tested to suppress artifacts and improve overall coherence. Finally, we analyze the impact of varying the guidance scale on the result. Multiple values are tested, and the resulting images are evaluated to determine the optimal balance between fidelity to the input and adherence to the learned priors of the diffusion model.

Based on the findings from these experiments, we identify the best-performing combination of parameters (as shown in \cref{final parameter set}), which is then adopted for the final test settings to ensure the optimal performance. 

\begin{table}[htbp]
    \centering
    \small
    \begin{tabular}{c|ccc}
    \toprule
        Datasets & Start point & gs & Text prompt \\
    \midrule
        synthetic & lr & 0.9 & w/o prompt \\
        wild & noise & 8.5 & w positive prompt \\
    \bottomrule
    \end{tabular}
    \caption{Hyperparameter selection for testing and validation. "gs" is the abbreviation of guidance scale.}
    \label{final parameter set}
\end{table}

\subsection{Quantitative Results}

\begin{table*}[htbp]
    \centering
    \begin{adjustbox}{valign=c, scale=0.88}
    \begin{tabular}{c|c|c@{\hspace{6pt}}c@{\hspace{6pt}}c@{\hspace{6pt}}c@{\hspace{6pt}}c@{\hspace{6pt}}c@{\hspace{6pt}}c@{\hspace{6pt}}c@{\hspace{6pt}}c@{\hspace{6pt}}c}
        \toprule
        Dataset & Metrics & BSRGAN\cite{BSRGAN} & Real-ESRGAN\cite{Real-ESRGAN} & FaithDiff\cite{faithdiff} & InvSR\cite{invsr} & XPSR\cite{xpsr} & PiSA-SR\cite{pisa} & SeeSR\cite{SeeSR} & Ours \\
        \midrule
        \multirow{5}{*}{synthetic} & MuSIQ$\uparrow$ & 70.5933 & 67.8211 & 72.5569 & 73.2262 & 69.8961 & \textcolor{red}{74.4527} & \textcolor{blue}{73.4914} & 69.7667 \\
        & ManIQA$\uparrow$ & 0.4090 & 0.4003 & 0.4344 & 0.4655 & \textcolor{red}{0.5266} & 0.5142 & \textcolor{blue}{0.5170} & 0.4523 \\
        & CLIPIQA$\uparrow$ & 0.6191 & 0.5781 & 0.6618 & 0.7287 & \textcolor{blue}{0.7449} & \textcolor{red}{0.7511} & 0.7295 & 0.6413 \\
        & NRQM$\uparrow$ & 6.3076 & 6.3502 & \textcolor{red}{6.6970} & \textcolor{blue}{6.6888} & 6.4588 & 6.6513 & 6.4621 & 6.1655 \\ 
        & HyperIQA$\uparrow$ & 0.5948 & 0.5599 & 0.6007 & 0.6299 & 0.6375 & \textcolor{red}{0.6649} & \textcolor{blue}{0.6593} & 0.6105 \\
        \midrule
        \multirow{5}{*}{wild} & MuSIQ$\uparrow$ & 61.8400 & 63.1264 & 66.1299 & 54.9506 & 62.4038 & \textcolor{blue}{70.3856} & 68.1723 & \textcolor{red}{71.0000} \\
        & ManIQA$\uparrow$ & 0.3811 & 0.4794 & 0.3947 & 0.3639 & 0.4321 & \textcolor{blue}{0.4980} & 0.4831 & \textcolor{red}{0.5529} \\
        & CLIPIQA$\uparrow$ & 0.5097 & 0.6553 & 0.6072 & 0.4897 & 0.6279 & \textcolor{blue}{0.7275} & 0.7024 & \textcolor{red}{0.7633} \\
        & NRQM$\uparrow$ & 6.2272 & 6.4495 & \textcolor{blue}{6.8256} & 5.3240 & 6.4639 & 6.7516 & 6.6068 & \textcolor{red}{6.9242} \\
        & HyperIQA$\uparrow$ & 0.5476 & 0.5901 & 0.5519 & 0.4967 & 0.5568 & \textcolor{blue}{0.6391} & 0.6237 & \textcolor{red}{0.6826} \\
        \midrule
        \multirow{3}{*}{DIV2K} & PSNR$\uparrow$ & \textcolor{blue}{23.3098} & \textcolor{red}{23.3343} & 22.5975 & 21.6333 & 21.4079 & 22.9385 & 22.7672 & 23.2955 \\
        & SSIM$\uparrow$ & 0.6237 & \textcolor{red}{0.6405} & 0.5931 & 0.5857 & 0.5526 & 0.6144 & 0.6044 & \textcolor{blue}{0.6240} \\
        \bottomrule
    \end{tabular}
    \end{adjustbox}
    \caption{Comparison of different SISR methods on synthetic, wild and DIV2K validation dataset. \textcolor{red}{Red} and \textcolor{blue}{blue} represent the best and second score, respectively.}
    \label{result_synthetic_wild}
\end{table*}

We compare our proposed method with several state-of-the-art SISR methods, including GAN-based methods BSRGAN \cite{BSRGAN}, Real-ESRGAN \cite{Real-ESRGAN} and diffusion-based methods Faithdiff \cite{faithdiff}, InvSR \cite{invsr}, XPSR \cite{xpsr}, PiSA-SR \cite{pisa}, SeeSR \cite{SeeSR}. The qualitative results are shown in \cref{result_synthetic_wild}.

On the synthetic validation dataset, although our model's performance is not outstanding in any particular metric, it consistently ranks at a mid-to-high level across all metrics, and overall, it performs well among these models.

On the wild validation dataset, our method consistently achieves the best performance across all metrics, surpassing previous SOTA methods and demonstrating its robustness in handling real-world degradations. 

On the DIV2K dataset, we compare our proposed method with other models using reference-based metrics. Our model ranks third in PSNR and second in SSIM. Although our model performs worse than REAL in objective metrics, it achieves the best performance in subjective evaluation.

This highlights the effectiveness of our semantic-guided diffusion framework in reconstructing fine details while preserving both structural integrity and perceptual consistency. Furthermore, these results validate the capability of our approach to bridge the gap between synthetic and real-world super-resolution scenarios, making it a promising solution for practical UGC image enhancement.

\subsection{Qualitative Results}
\begin{figure*}[htbp]
    \centering
    \includegraphics[width=0.8\linewidth]{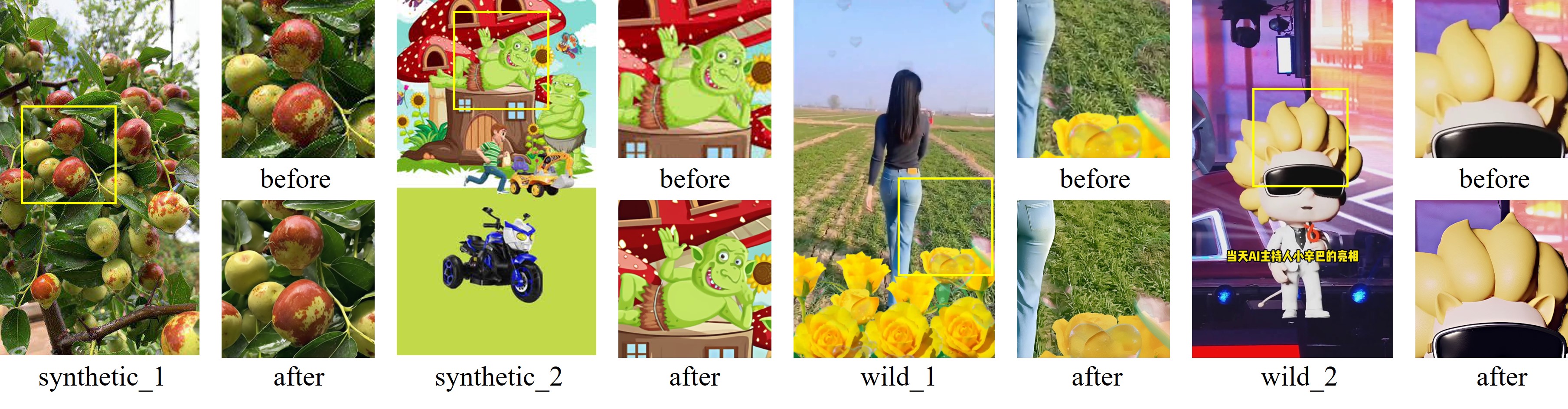}
    \caption{Comparison of images from the wild and synthetic datasets before and after 1$\times$ and 4$\times$ super-resolution processing with our model.}
    \label{fig:process_cmp}
\end{figure*}

The images before and after the restoration process are presented in \cref{fig:process_cmp}. As illustrated in \cref{fig:process_cmp}, the proposed model demonstrates a strong capability in reconstructing fine details within low-resolution images. Furthermore, the restored images exhibit improved visual clarity and structural integrity, contributing to a more realistic and perceptually pleasing outcome. This suggests that the model not only enhances objective image quality but also maintains high subjective fidelity, making it suitable for applications requiring both detail preservation and natural visual appearance.

\begin{figure*}[htbp]
    \centering
    \begin{subfigure}{\textwidth}
        \centering
        \includegraphics[width=0.9\linewidth]{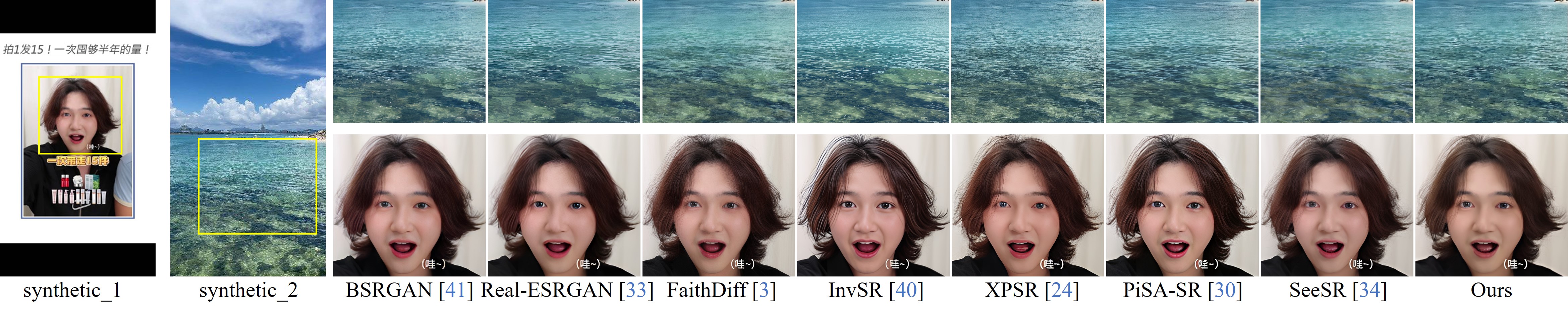}
    \end{subfigure}

    \begin{subfigure}{\textwidth}
        \centering
        \includegraphics[width=0.9\linewidth]{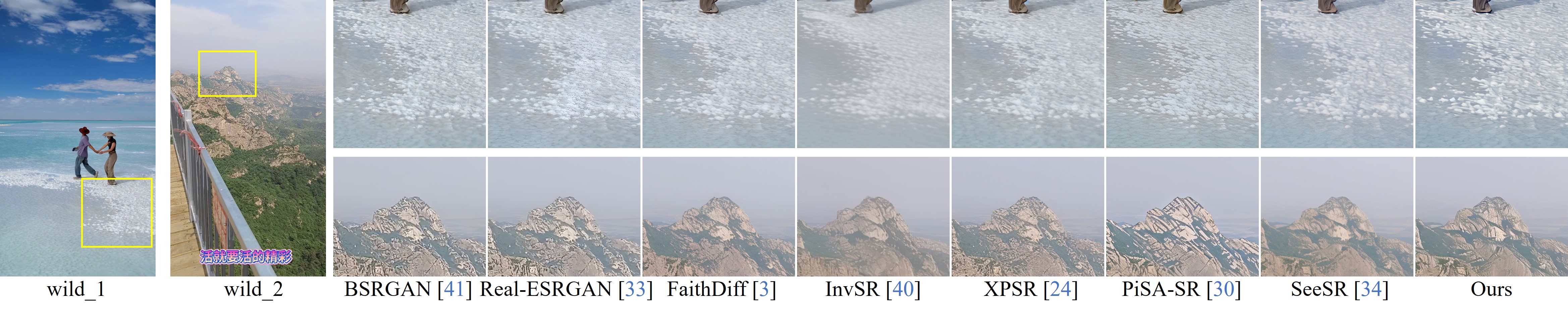} 
    \end{subfigure}
    
    \caption{Comparison of images with different models on synthetic and wild validation dataset.}
    \label{fig:s_w_model_cmp}
\end{figure*}

We also compared the perceived realism of the proposed method with existing state-of-the-art methods on wild and synthetic data (as illustrated in Figures \cref{fig:s_w_model_cmp}). The images produced by our proposed method exhibit a higher degree of naturalness and realism compared to those generated by other approaches based on Generative Adversarial Networks (BSRGAN \cite{BSRGAN}, Real-ESRGAN \cite{Real-ESRGAN}) or Diffusion models (FaithDiff \cite{faithdiff}, InvSR \cite{invsr}, XPSR \cite{xpsr}, PiSA-SR \cite{pisa}, SeeSR \cite{SeeSR}). This demonstrates the effectiveness of our method in synthesizing visually coherent and perceptually convincing results.

\subsection{Ablation Study}
\noindent\textbf{Text Prompt}
To evaluate the impact of text prompts on model performance, we conducted ablation experiments by introducing additional positive and negative prompts during inference. Results across various guidance scale settings indicate that both positive and negative prompts contribute to the final output in a comparable manner.

As shown in \cref{prompt_score}, for the wild dataset, incorporating additional positive prompts yields the highest scores, suggesting an improvement in model performance. In contrast, for the synthetic dataset, prompts have minimal effect on enhancing results.

Based on these observations, our inference strategy is as follows: for the wild dataset, we include additional positive prompts to refine outputs, whereas for the synthetic dataset, no additional prompts are introduced, as they do not contribute to further improvements.

\begin{table}[htbp]
    \centering
    \small
    \begin{tabular}{c|cc}
    \toprule
        Dataset & Text prompt  & Score$\uparrow$ \\
    \midrule
       \multirow{3}{*}{wild} & w/o prompt & $20.2182$ \\
        & w positive prompt & $\mathbf{20.5054}$ \\
        & w negative prompt & $19.7869$ \\
    \midrule
       \multirow{3}{*}{synthetic} & w/o prompt & $\mathbf{33.2336}$ \\
        & w positive prompt & $33.2312$ \\
        & w negative prompt & $\mathbf{33.2336}$ \\
    \bottomrule
    \end{tabular}
    \caption{Impact of additional text prompts on model performance across both datasets. Positive prompt includes "clean, high-resolution, 8k, ultra-detailed, ultra-realistic". Negative prompt includes "dotted, noise, blur, lowres, smooth, unrealistic physics, unnatural shadows". For wild dataset, score is computed as: $\text{Score}=0.1\times \text{MUSIQ}+10\times \text{ManIQA}+10\times \text{CLIPIQA}$. For synthetic dataset, score is computed as: $\text{Score}=\text{PSNR}+10\times \text{SSIM}-10\times \text{LPIPS}$}
    \label{prompt_score}
\end{table}

\noindent\textbf{Guidance Scale}
We evaluated the final scores of the generated results across different guidance scale values, as shown in \cref{gs_score}. For the synthetic dataset, the highest score is achieved when the guidance scale is set to 0.9, indicating that this setting optimally enhances performance in this scenario. In contrast, for the wild dataset, there is a positive correlation between the guidance scale and the final evaluation score.

\begin{table}[htbp]
    \centering
    \small
    \begin{tabular}{ccc|ccc}
    \toprule
        Dataset & gs  & Score$\uparrow$ & Dataset & gs  & Score$\uparrow$ \\
    \midrule
       \multirow{4}{*}{wild} & $8.5$ & $20.2305$ & \multirow{4}{*}{synthetic} & $1.1$ & $33.2196$ \\
        & $10$ & $20.3897$ & & $1.0$ & $33.2335$ \\
        & $12$ & $20.4619$ & & $0.9$ & $\mathbf{33.2340}$\\
        & $14$ & $\mathbf{20.5054}$ & & $0.8$ & $33.2338$ \\
    \bottomrule
    \end{tabular}
    \caption{Impact of guidance scale on wild and synthetic dataset. "gs" is the abbreviation of guidance scale }
    \label{gs_score}
\end{table}

\begin{figure*}[htbp]
  \centering
  \includegraphics[width=0.72\linewidth] {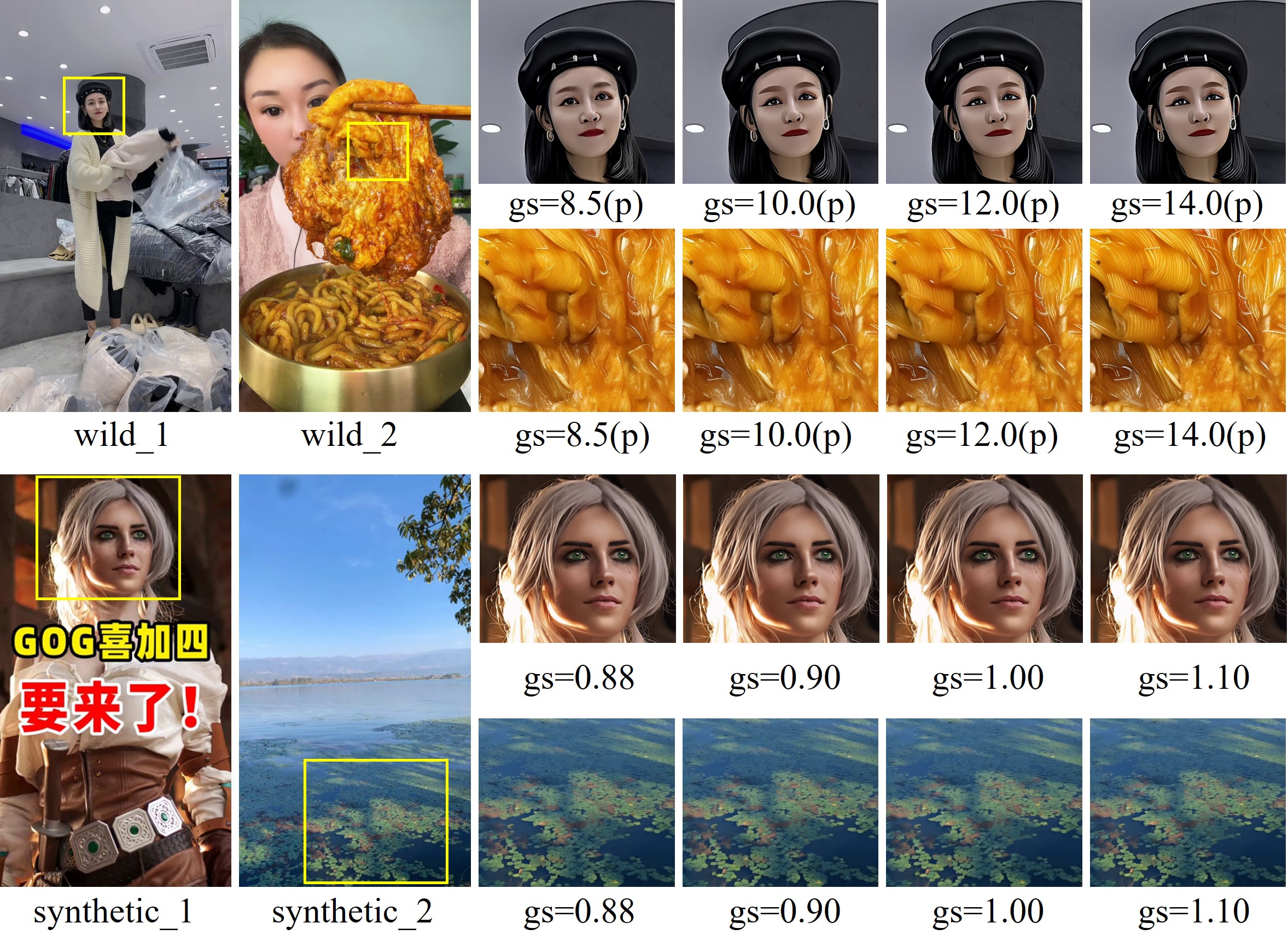}
  \caption{Image restoration results of wild and synthetic datasets under different gs values.}
  \label{fig:gs}
\end{figure*}

To further analyze the impact of guidance scale on image perceptual quality, we examine the results for the wild test set at values of 8.5, 10, 12, and 14, and for the synthetic test set at values of 0.88, 0.9, 1 and 1.1, as illustrated in \cref{fig:gs}. For the synthetic dataset, when the guidance scale is set around 0.9, the generated images exhibit minimal perceptual differences. For the wild dataset, increasing the guidance scale generally enhances the evaluation score. However, beyond a certain threshold, the generated images exhibit noticeable artifacts, resulting in a degradation of perceptual quality. To balance evaluation metrics and perceptual quality, we set the guidance scale to 8.5 for the wild test set and 0.9 for the synthetic test set.

\begin{figure*}[htbp]
    \centering
    \includegraphics[width=0.81\linewidth]{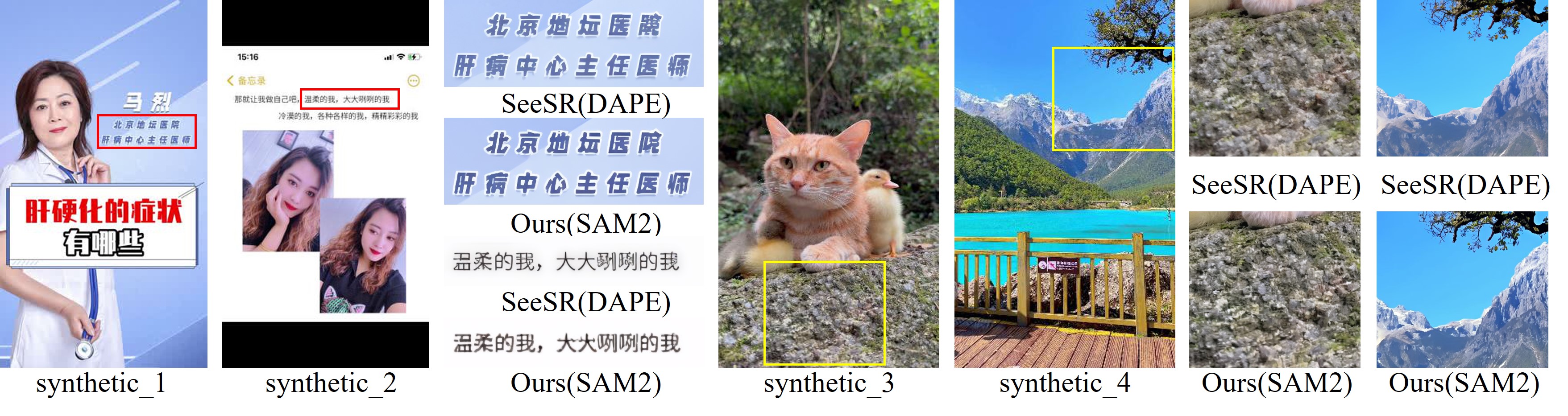}
    \caption{Comparison of the effects of the DAPE \cite{SeeSR} and SAM2 \cite{SAM2} modules on synthetic dataset. The results of two models are obtained under the same parameter setting (gs=$0.9$ withou extra positive prompt).}
    \label{fig:sam2_cmp}
\end{figure*}

\noindent\textbf{Semantic-Aware Module}
To evaluate the effectiveness of the proposed semantic-aware module, we compare it with the DAPE module used in SeeSR \cite{SeeSR}. To ensure a fair comparison, we standardize the parameter settings for both models and conduct qualitative and quantitative analyses to assess their performance.

For the wild dataset, we evaluate performance using the MUSIQ, MANIQA, and CLIPIQA metrics. As shown in \cref{sematic_comp_wild}, our semantic-aware module outperforms DAPE across all metrics, indicating that the SAM2 \cite{SAM2} module enhances the model's ability to process wild images compared to the DAPE module. For the synthetic dataset, due to the lack of GT images, we focus on subjective evaluations, as shown in \cref{fig:sam2_cmp}. As the example shown in \cref{fig:sam2_cmp}, our model demonstrates superior capability in restoring text details and natural texture compared to SeeSR. Overall, the results suggest that our proposed semantic-aware module significantly improves performance on both the wild and synthetic datasets.

\begin{table}[htbp]
    \centering
    \small
    \begin{tabular}{{c|c@{\hspace{6pt}}c@{\hspace{6pt}}c@{\hspace{6pt}}c}}
    \toprule
        Model & MUSIQ$\uparrow$ & MANIQA$\uparrow$ & CLIPIQA$\uparrow$ & Score$\uparrow$ \\
    \midrule
       DAPE & $70.3434$ & $0.5332$ & $0.7345$ & $19.7119$ \\
       Ours & $\mathbf{71.1969}$ & $\mathbf{0.5532}$ & $\mathbf{0.7579}$ & $\mathbf{20.2305}$ \\
    \bottomrule
    \end{tabular}
    \caption{Ablation study of semantic-aware model on wild dataset.}
    \label{sematic_comp_wild}
\end{table}

\section{Conclusion}
\label{sec:conclusion}
In this paper, we propose a novel approach to UGC image super-resolution by incorporating semantic guidance into a diffusion-based framework. By leveraging semantic embeddings, our method enriches the model with high-level contextual information, enabling more accurate reconstruction of fine details while mitigating artifacts. To address the disparity between synthetic and real-world degradations, we separately simulate both types on the LSDIR dataset and integrate them with the official training set, resulting in a more diverse and representative dataset. Additionally, we utilize the pretrained SAM2 model to extract refined structural and semantic features, further enhancing the quality of super-resolved images. Through meticulous selection of key hyperparameters, we optimize the model for perceptual fidelity, ensuring visually coherent and high-quality results. Our approach effectively bridges the gap between synthetic and real-world super-resolution, contributing to more robust, perceptually faithful, and practically applicable image enhancement.

Despite the promising results, our method still presents several limitations. First, it exhibits notable difficulties in reconstructing text regions under conditions of severe degradation or occlusion. Secondly, in regions with substantial corruption or low resolution, our method may inadvertently introduce excessive high-frequency components, potentially leading to a degradation in the subjective perceptual quality. While the current approach has certain constrains, it provides a strong foundation for future extensions in this domain.

\section*{Acknowledgements}
\label{sec:acknowledgements}
This work was partly supported by Science and Technology Commission of Shanghai Municipality (No. 24511106200), the Shanghai Key Laboratory of Digital Media Processing and Transmission under Grant 22DZ2229005, 111 project BP0719010.

{
    \small
    \bibliographystyle{ieeenat_fullname}
    \bibliography{main}
}


\end{document}